\documentclass{ifacconf}

\usepackage{graphicx}      
\usepackage{natbib}        
\usepackage{algorithm}
\usepackage{algpseudocode}
\usepackage{amsmath}
\usepackage{amssymb}
\usepackage{subcaption}
\begin{document}
\begin{frontmatter}

\title{Low Resolution Next Best View for Robot Packing} 

\author[First]{Giuseppe Fabio Preziosa} 
\author[First]{Chiara Castellano} 
\author[First]{Andrea Maria Zanchettin}
\author[First]{Marco Faroni}
\author[First]{Paolo Rocco}

\address[First]{The Authors are with Politecnico di Milano, Dipartimento di Elettronica,Informazione e Bioingegneria (DEIB), Piazza Leonardo da Vinci 32, 20133, Milano (Italy). (e-mail: giuseppefabio.preziosa, chiara2.castellano, andreamaria.zanchettin, marco.faroni, paolo.rocco)@polimi.it}

\begin{abstract}                
Automating the packing of objects with robots is a key challenge in industrial automation, where efficient object perception plays a fundamental role. This paper focuses on scenarios where precise 3D reconstruction is not required, prioritizing cost-effective and scalable solutions. The proposed Low-Resolution Next Best View (LR-NBV) algorithm leverages a utility function that balances pose redundancy and acquisition density, ensuring efficient object reconstruction. Experimental validation demonstrates that LR-NBV consistently outperforms standard NBV approaches, achieving comparable accuracy with significantly fewer poses. This method proves highly suitable for applications requiring efficiency, scalability, and adaptability without relying on high-precision sensing.
\end{abstract}


\end{frontmatter}

\section{Introduction}

Automation in packaging and warehouse logistics has emerged as a pivotal area of industrial research, driven by the increasing demand for flexible and customizable production. Among the many processes impacted by this shift, the final packaging phase—where objects are identified, their dimensions determined, and placed in appropriately sized containers—represents a shared challenge that spans industries and product types. Currently, these tasks are predominantly performed by human workers, with container selection and object arrangement often relying on intuition and manual effort. This dependency introduces inconsistencies and limits scalability, particularly in dynamic, high-volume environments.

Research in robotic packing has traditionally focused on optimizing spatial efficiency within packages. \cite{Packing_just_ordering_offline} tackled the challenge of optimizing space while considering package stability, whereas more recently, \cite{Packing_just_ordering} focused on exploring optimal spatial arrangements to maximize efficiency with both known and unknown objects. These methods often rely on high-resolution sensors and precise planning designed to be implemented offline before the arrival of objects. However, a shift has emerged towards solutions based on low-cost and easily scalable hardware. These approaches explore simpler sensing systems, such as RGB cameras in \cite{Packing_rgb_data}, and minimalistic grippers with \cite{Packing_low_cost}, aiming to provide practical and adaptable solutions for real-world applications. 
Along this line, the present work addresses the perception challenges associated with the use of \emph{low-resolution sensing} systems in robot packing processes.

\begin{figure}
\begin{center}
\includegraphics[width=8cm]{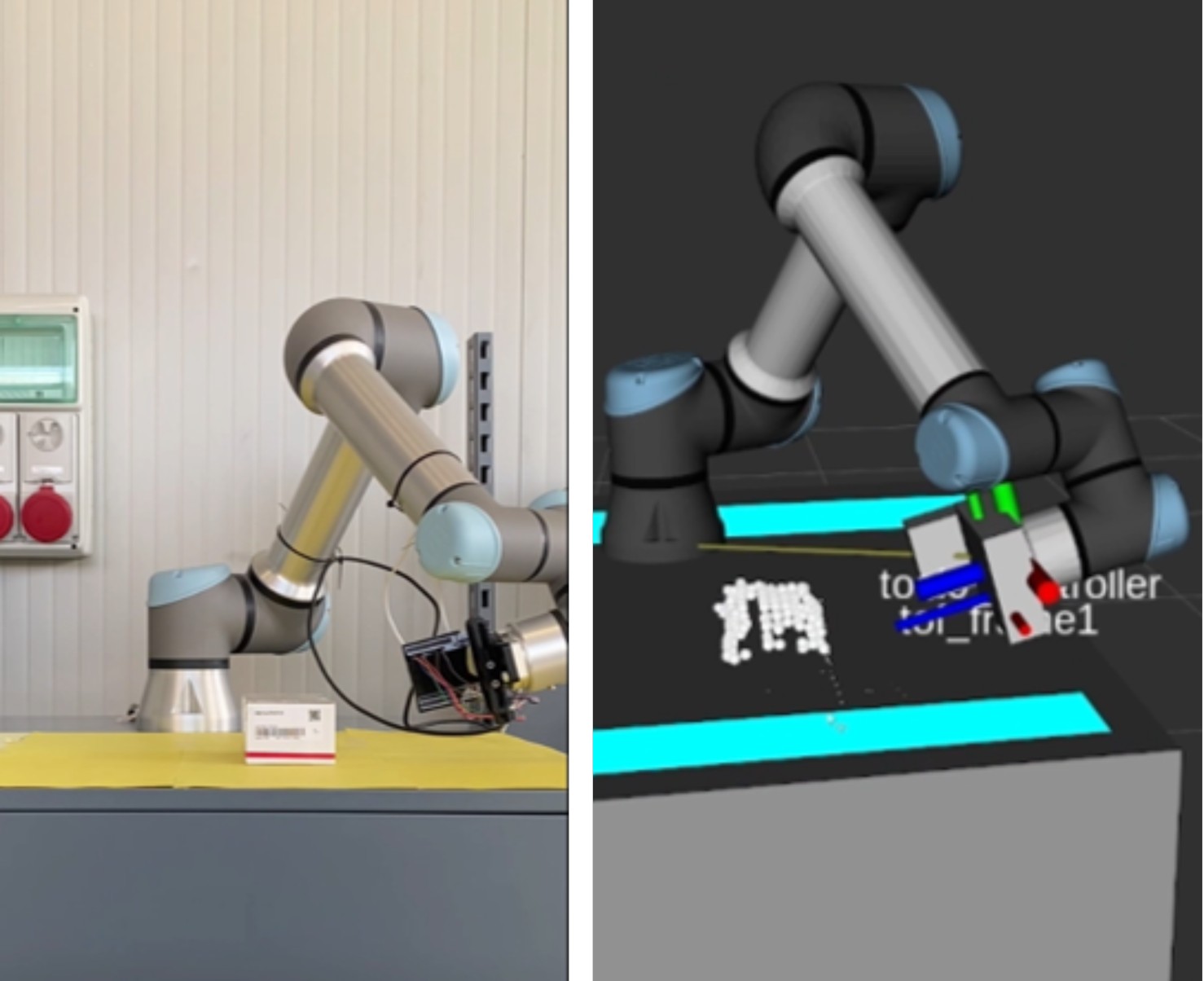}    
\caption{Last acquisition of the LR-NBV algorithm showing on the right the final result of the reconstruction. } 
\label{fig:nbv_box}
\vspace{-0.3cm}
\end{center}
\end{figure}

Despite the proven effectiveness of low-resolution sensors in other contexts, such as human-robot safety, as demonstrated by \cite{spots}, their use for object reconstruction presents significant challenges. Their reduced fidelity can make it difficult to capture sufficient detail with a single acquisition, often requiring additional viewpoints to build an adequate representation of the object. In packaging tasks, however, detailed 3D reconstructions are not strictly necessary. Instead, efficiently estimating object dimensions and bounding volumes is sufficient for subsequent steps, such as grasp planning, container selection and arrangement of the objects. Studies like \cite{graspingPaper1, graspingPaper2, graspingPaper3} demonstrate that even partial reconstructions, covering only 25–30\% of an object surface, can enable successful robotic manipulation. This highlights how low-resolution sensors can effectively balance efficiency and precision, offering scalability and cost advantages.

However, low-resolution sensing introduces further challenges, such as reducing the number of acquisitions while managing viewpoint diversity evaluation. Traditional \emph{Next Best View (NBV)} algorithms, designed for high-resolution sensors, often fail to adapt to these constraints, necessitating the development of tailored solutions. To address these limitations, this paper proposes a novel Low-Resolution Next Best View (LR-NBV) algorithm, which, to the best of our knowledge, is one of the first works to introduce and explore the use of low-cost and easily integrable sensors for perception tasks like this.

\section{Related Works}

The problem of observing and evaluating the best next viewpoint to gain information about the environment is a well-known challenge, referred to as the \emph{Next Best View} problem. Initially introduced in the early 1990s by~\cite{old_1, old_2, base_NBV}, this concept has been extensively studied and applied to tasks such as exploration, object reconstruction, and inspection, across a wide range of robotic applications, including both mobile robots and fixed manipulators. The characterization of an NBV algorithm involves addressing several core aspects, which provide a clear framework for classifying our proposed approach. These aspects include task definition, spatial representation, observation evaluation, utility functions, constraints, and planning strategies.

NBV algorithms can be broadly categorized into \emph{model-based view planning} and \emph{non-model-based view planning}. As outlined in~\cite{base_NBV}, model-based approaches aim to verify an existing model by comparing it to a physical object, often for defect detection or conformity checks. In contrast, non-model-based approaches, which are the focus of this work, assume no prior knowledge of the object and aim to incrementally build its representation from scratch. This distinction is particularly relevant for 3D object reconstruction in unstructured environments, where the lack of prior information necessitates flexible and efficient strategies. Non-model-based methods often rely on volumetric representations to discretize the environment into manageable units. These methods segment the space into voxel grids, typically combined with an associated probability and hierarchical structure, as represented by the octomap framework in \cite{octomap}. Each voxel is classified as free, occupied, or uncertain, simplifying visibility operations and enabling probabilistic updates of voxel states based on sensor data. Viewpoint evaluation is typically conducted using raycasting, where rays are cast from the sensor into the voxelized space, simulating the sampling process of a camera. As the rays traverse the voxel grid, each voxel state is updated probabilistically, enabling the estimation of information gain.

\cite{sampling_volumetric} highlighted how the combination of volumetric representations and sampling-based algorithms provides a flexible framework for defining utility functions. These utility functions, which are at the core of NBV algorithms, guide the selection of viewpoints by quantifying their informativeness and can be tailored to specific tasks and hardware constraints, providing the foundation for effective exploration strategies. Classic NBV algorithms typically define utility functions based entirely on exploration gain, which focuses on maximizing the amount of new information acquired. For instance, \cite{1_utility} introduce an AreaFactor that balances the observation of different voxel types, while \cite{2_utility_volumetric} propose a volumetric gain that is computationally efficient, making it both fast and suitable for a packing task. Additionally, \cite{3_utility} directly counts the number of unknown voxels along the raycast path to quantify exploration. Finally, \cite{5_utility} presents multiple formulations for exploration gain, including metrics such as occlusion-aware and rear-side voxel information, optimized for volumetric 3D reconstruction. On the other hand, fewer studies incorporate hardware-specific considerations into utility functions, often referred to as \emph{quality gains}. For example, \cite{1_utility} prioritizes viewpoints that minimize robot motion by favoring poses requiring less movement. Similarly, in sensor-specific scenarios, \cite{4_utility} optimize the sensor acquisition angle, with \cite{4_utility} positively weighting viewpoints that are perpendicular to the surface.

Although effective, the quality gains discussed earlier, when combined with classical exploration gains, face limitations in addressing the challenges posed by low-resolution sensors. A key limitation of these sensors is their inability to differentiate effectively between acquisitions made from slightly different positions. Due to the sparsity of the data describing the object, traditional utility functions assign similar exploration weights to these redundant viewpoints, leading to unnecessary acquisitions. In robotic packing, where speed and efficiency are paramount, this redundancy significantly slows down the process. To address redundancy, \cite{reference_paper} introduces a \emph{Visited Gain} that penalizes redundant acquisitions by negatively weighting revisited areas. While this gain was designed for mobile manipulators, our work extends its application to fixed manipulators. Furthermore, we introduce additional gains tailored to the unique constraints of low-resolution sensors, aiming to optimize the acquisition process and enhance efficiency in robotic packing tasks. 

Additionally, this work differs from many approaches requiring prior knowledge of the object position and size. For instance, \cite{sphere_1}, and \cite{sphere_3} sample viewpoints on a sphere centered around the object, which implicitly requires the knowledge of the object pose and dimension. In contrast, this work structures sampling directly within the 3D space of the sensor pose, using only the robot workspace boundaries as constraints. Similar to \cite{no_assump_only_first_acqu}, the algorithm requires only an initial acquisition of the object to start the process, allowing it to adapt seamlessly to objects of varying sizes and positions.

This work also belongs to the family of approaches that expand a tree of viewpoints in the sampling space, drawing inspiration from RRT-based methods. At each iteration, the tree is incrementally built starting from the end effector pose and integrates updated information from the environment. This ensures that the planning process adapts dynamically as new data are acquired. Unlike classical RRT, where the entire path is planned upfront, this algorithm follows the receding horizon paradigm introduced in \cite{receding_horizon}. In this approach, the tree is constructed iteratively, and only the first segment of the most promising branch—evaluated based on utility gain —is executed. The process is then repeated, progressively refining the tree and adapting to the newly acquired map of the environment. To address the risk of getting trapped in local minima, as highlighted by \cite{2_utility_volumetric} and \cite{reference_paper}, the algorithm retains the most promising nodes from previous iterations, reevaluating them in subsequent cycles. This strategy reduces computational costs while providing useful indices for monitoring reconstruction progress and defining termination criteria, enhancing the robustness of the exploration process.

\section{Proposed Approach}

\begin{algorithm}
\caption{Low Resolution Next Best View Algorithm}
\label{alg:nbv}
\begin{algorithmic}[1]

\State {\textbf{Inputs:} $A^0, C^0, n$ given.}
\State $\text{completed} \gets \text{False}$
\State $i \gets 0$
\State $B^{*i} \gets \emptyset$
\State $O \gets \emptyset$

\While{$\neg completed$}
    \State $(B^{*i}, C_{\mathcal{T}^i}) \gets \text{ExpandRRT}(n, A^i)$
    \State $C^i \gets \text{AppendNodes}(C^{i-1}, C_{\mathcal{T}^i})$
    \State $x_T^* \gets \arg\max_{x_T \in B^{*i}} U(x_T)$
    \If{$U(x_T^*) < U(x_T^{\text{bestC}})$}
        \State $(B^{*i}, C_{\hat{\mathcal{T}}^i}) \gets \text{ExpandRRTCached}(C^i, A^i)$
        \State $C^i \gets \text{AppendNodes}(C^i, C_{\hat{\mathcal{T}}^i})$
        \State $x_T^* \gets \arg\max_{x_T \in B^{*i}} U(x_T)$
    \EndIf
    \If{$U(x_T^*) > U(x_T^{\text{worstC}})$}
        \State $D^i \gets \text{captureData}(B^{*i})$
        \State $A^i \gets \text{UpdateOctomap}(D^i, A^{i-1})$
        \State $O^i \gets \text{ExtractObjectPoints}(A^i)$
        \State $C^i \gets \text{UpdateCachedNodes}(C^{i-1}, A^i)$
        \State $x_T^{\text{bestC}} \gets \arg\max_{x_T \in C^i} U(x_T)$
        \State $x_T^{\text{worstC}} \gets \arg\min_{x_T \in C^i} U(x_T)$
        \State $i \gets i + 1$
    \Else
        \State $completed \gets \text{True}$
    \EndIf
\EndWhile

\State{\textbf{return} $O^i$}

\end{algorithmic}
\end{algorithm}

The approach presented in this paper aims to identify an efficient sequence of sensor poses, reducing the makespan of the acquisition while ensuring a sufficient reconstruction of the object surface to enable the robot to correctly grasp the object. Each sensor pose \( x_T \in SE(3) \) generates an observation \( D(x_T) \), defined as a set of \( M \) points \( \{p_1, p_2, \ldots, p_M\} \subseteq \mathbb{R}^3 \), where \( M \) represents the resolution of the sensor and determines the number of points captured during each acquisition. The object \( O \) reconstructed is a function of the sequence of observations collected from the \( k \) poses selected by the algorithm, formally expressed as \( O = f(D(x_{T_1}), D(x_{T_2}), \ldots, D(x_{T_k})) \). To guide the selection of these poses, the algorithm employs a utility function \( U(x_{T_j}) \), which evaluates the expected contribution of each pose to the reconstruction process. This function balances the need to explore new areas of the object and avoid redundant measurements caused by the sensor limited resolution.

The three main components of this algorithm are presented in the following sections. In particular, Section~\ref{section::Sampling-Based RRT} describes the pose-searching algorithm, which serves as the backbone of the entire approach. Section~\ref{Object Clustering and Dynamic ROI} explains how the object is identified and characterized based on data extracted from the environment after observations. Finally, Section~\ref{utility function} provides a detailed explanation of the utility function used during the evaluation process.



\subsection{Sampling-Based RRT}
\label{section::Sampling-Based RRT}

The proposed algorithm follows a cyclic structure inspired by state-of-the-art approaches described by~\cite{receding_horizon, reference_paper}. At each iteration \( i \), starting from line 7 in Algorithm \ref{alg:nbv}, a tree \( \mathcal{T}^i \) is constructed through the expansion step. Specifically, the tree \( \mathcal{T}^i \) is initialized from the current sensor pose, which is determined by the robot end-effector pose, and its branches are expanded by sampling  \( n \) points in the 3D space constrained by the robot workspace. For each sampled point, a corresponding pose \( x_T' \in SE(3) \) is generated, where the orientation is determined to maximize the utility function \( U(x_T') \). From this tree, the best branch \( B^{*i} \) is selected by maximizing the cumulative utility of its nodes:
\begin{equation} \label{eq:best_branch}
B^{*i} = \mathop{\mbox{argmax}}_{B_k^i \in \mathcal{T}^i} \sum_{x_T \in B_k^i} U(x_T)
\end{equation}
From the best branch, the best node is then extracted (line 9). Finally, the best branch is used to suggest the next poses for data acquisition (line 16).
The algorithm also evaluates the remaining branches of \( \mathcal{T}^i \), identifying nodes not included in \( B^{*i} \) but with utility \( U(x_T) \) exceeding a threshold \( \tau \). These nodes, referred to as \emph{cached nodes}, are stored in \(  C_{{\mathcal{T}}^i} \) and added to the global structure \(  C{^i} \) (lines 8 and 12). Since a new tree is created at every iteration, cached nodes play a crucial role in mitigating issues related to local minima. Specifically, the cached nodes from the previous iteration \( i-1 \) are analyzed to extract the best and worst nodes according to their utility. The \emph{best cached node} is used to decide whether it is worth expanding the tree toward the cached nodes (lines 10 and 11), where the tree is expanded by sampling from the most promising cached nodes. On the other hand, the \emph{worst cached node} provides a utility gain threshold to determine whether exploration should continue. The exploration is considered complete if the utility of the best node falls below the utility of the worst cached node (line 24).
It is also important to note that at the end of each iteration, the cached nodes are updated using new data acquired during the process. This update reorders the cached nodes, keeping only a predefined number of nodes \( m \), which decreases linearly as iterations progress. This mechanism ensures that only the most relevant nodes are retained, improving the efficiency of subsequent iterations.


\subsection{Object Clustering and Dynamic ROI}
\label{Object Clustering and Dynamic ROI}
One of the key strength of this algorithm is that no prior knowledge of the object pose and dimensions are required. By integrating sensor measurements into the volumetric octomap \( A \), the algorithm dynamically updates voxel occupancy probabilities, categorizing them as occupied, uncertain, or empty (line 17 in Algorithm \ref{alg:nbv}). This process enables efficient segmentation of the environment, laying the groundwork for object identification. To achieve this, the algorithm applies the Density-Based Spatial Clustering of Applications with Noise (DBSCAN) from~\cite{DBSCAN}, which identifies clusters based solely on point density. Its adaptability to irregular shapes and independence from prior object information make it computationally efficient for this task (line 18 in Algorithm \ref{alg:nbv}). Depending on the type of sensor used and insights gained through adequate sensor characterization, additional filters may need to be applied to DBSCAN to refine the clustering process and handle sensor-specific noise effectively.
Once the object \( O^i \) is identified, it is enclosed within a bounding box that serves two key roles. First, it defines the \emph{Region of Interest (ROI)}, guiding the utility function to prioritize reconstruction efforts in relevant regions while avoiding unnecessary exploration. Second, it acts as a \emph{Collision Checking} element, facilitating safe tree node generation and robot motion planning.

\begin{figure}
\begin{center}
\includegraphics[width=8.3cm]{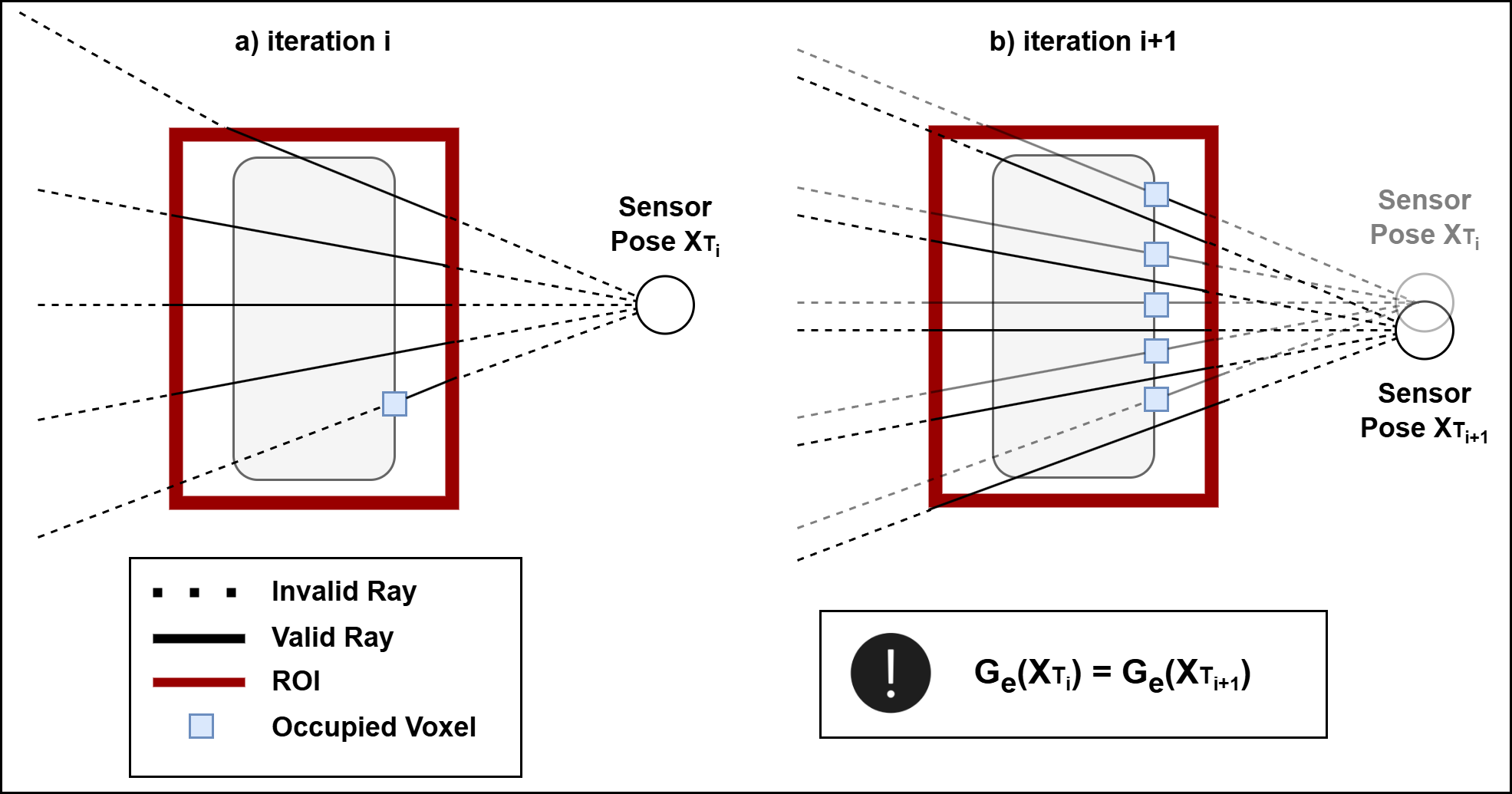}    
\caption{Validity of rays and exploration gains are shown in two subsequent iterations of the algorithm. Relying solely on exploration gain makes pose selection challenging, as poses slightly different from the previous ones yield the exact same gain.} 
\label{fig:ray_validity}
\end{center}
\end{figure}

\subsection{Utility Function}
\label{utility function}

The proposed utility function evaluates the potential of a candidate pose \( x_T \) by combining four main components: Exploration Gain (\( G_e \)), Density Gain (\( G_d \)), Quality Gain (\( G_q \)), and Visited Gain (\( G_v \)).
The utility function, therefore, takes the form of:

\vspace{-1.2em}
\[
U(x_n) = w_eG_e(x_T) + w_dG_d(x_T) + w_qG_q(x_T) + w_vG_v(x_T)
\]
\vspace{-1.2em}

where \( w_e, w_d, w_q, w_v >0 \) are the weights associated with each gain and can be adjusted to balance the relative importance of each component based on the specific task.

\subsubsection{Exploration Gain (\( G_e \)) and Density Gain (\( G_d \))} \label{exploration_gain}

The first two gains are based on a raycasting process that simulates the sensor behavior at a given pose. This process is defined by the sensor intrinsic properties, such as field of view (FoV) and resolution (res), which determine the spatial distribution of the emitted rays. Each ray \( \bold{r} \in \mathcal{R}(x_n, \text{FoV}, \text{res}) \) originates from the sensor position \( x_T \) and can be defined as: \( \bold{r}(t) = \mathbf{o} + t \mathbf{d} \),
where \( \mathbf{o} \in \mathbb{R}^3 \) is the ray origin, \( \mathbf{d} \in \mathbb{R}^3 \) is its direction, and \( t \) represents the distance along the ray. We define \( t^* \) as the first point along the ray \( \bold{r}(t) \) that intersects the object \( O^i \), represented as a set of points grouped as described in Section~\ref{Object Clustering and Dynamic ROI}. Based on this, we introduce the concept of ray validity as:
\begin{small}
\begin{equation} \label{eq:ray_validity}
V(\bold{r}, t) =
\begin{cases} 
1 & \text{if } \bold{r}(t) \cap \text{ROI} \neq \emptyset \text{ and } \exists t^* \leq t : \bold{r}(t^*) \in O^i, \\
0 & \text{otherwise.}
\end{cases}
\end{equation}
\end{small}
where, as visually explained in Fig.~\ref{fig:ray_validity}a, a portion of ray is considered valid only if it lies within the ROI and respects the occlusion constraint. This approach simulates the impossibility of perceiving regions beyond an occlusion represented by the object, ensuring that only the visible portions of the workspace contribute to the reconstruction process.
Based on that, the \emph{Exploration Gain \( G_e(x_T) \)} is defined as:
\begin{equation} \label{eq:exploration_gain}
G_e(x_T) = \sum_{\bold{r} \in \mathcal{R}(x_T, \text{FoV}, \text{res})} \sum_{t_k = 0}^{1} V(\bold{r}, t_k)W(\bold{r}, t_k)
\end{equation}
where \( t_k \) represents discrete samples along the ray, uniformly spaced between \( t = 0 \) and \( t = 1 \). In this formulation, \( W(\bold{r}, t_k) \) quantifies the contribution of each ray segment to volumetric exploration, accounting for the volume traversed in voxels as described by~\cite{2_utility_volumetric}. The combination of ray validity and volumetric exploration focuses the process on areas of high uncertainty, maximizing the acquisition of new information.

However, the use of a purely exploratory gain does not adequately weigh or prioritize one pose over another, especially in the context of low-resolution sensors. As illustrated in Fig.\ref{fig:ray_validity}b, the same \( G_e \) value can be associated with two slightly different poses, as the rays would be entirely unobstructed and thus considered fully explorable. This limitation arises because \( G_e \) does not account for the density of object points in the explored regions, potentially leading to redundant acquisitions.
To address this issue, we introduce the \emph{Density Gain \( G_d \)}, designed to complement \( G_e \) by favoring regions with sparse object points, as:
\begin{equation} \label{eq:density_gain}
G_d(x_T) = \sum_{\bold{r} \in \mathcal{R}(x_T, \text{FoV}, \text{res})} \sum_{t_k = 0}^{1} V(\bold{r}, t_k) \cdot \frac{1}{1 +\rho(\bold{r}, t_k)}
\end{equation}
Here,  \( \rho(\bold{r}, t_k) \) represents the number of object points within a spherical neighborhood of radius \( r_d \) centered on \( \bold{r}(t_k) \).
This approach prioritizes regions with lower point density by penalizing rays in densely populated areas reducing redundancy.

\subsubsection{Quality Gain (\( G_q \)) and Visited Gain (\( G_v \))} \label{quality_gain}

The \emph{Quality Gain \( G_q \)} is particularly important when working with low-resolution and low-cost sensors, as it prioritizes acquisitions closer to the object to reduce noise. It is defined as:
\begin{equation} \label{eq:quality_gain}
G_q(x_T) = \frac{1}{1 +\|\ x_T - x_{\text{object}} \|}
\end{equation}
where \( x_{\text{object}} \) represents the estimated position of the object center. This formulation ensures that closer poses receive higher weights, promoting lower-noise acquisitions.

\begin{figure*}[h]
\begin{center}
\begin{tabular}{cc}
{\hspace{-0.4cm}\includegraphics[width=0.5\textwidth]{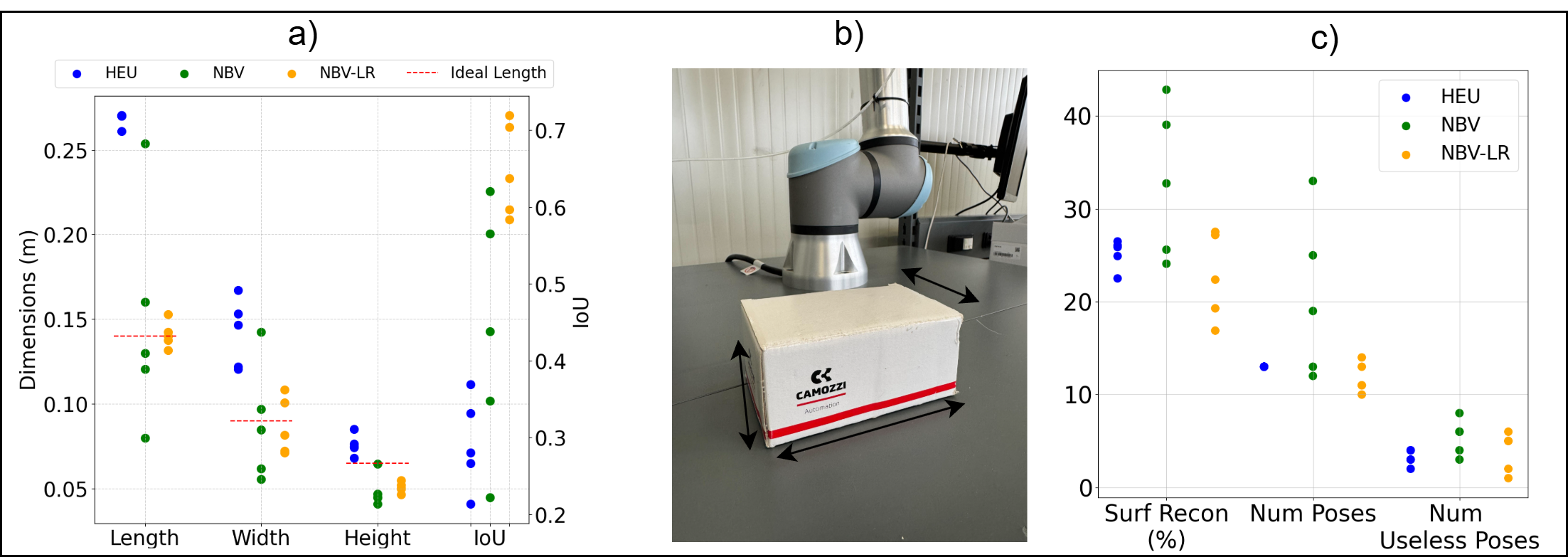}} & 
{\hspace{-0.1cm}\includegraphics[width=0.5\textwidth]{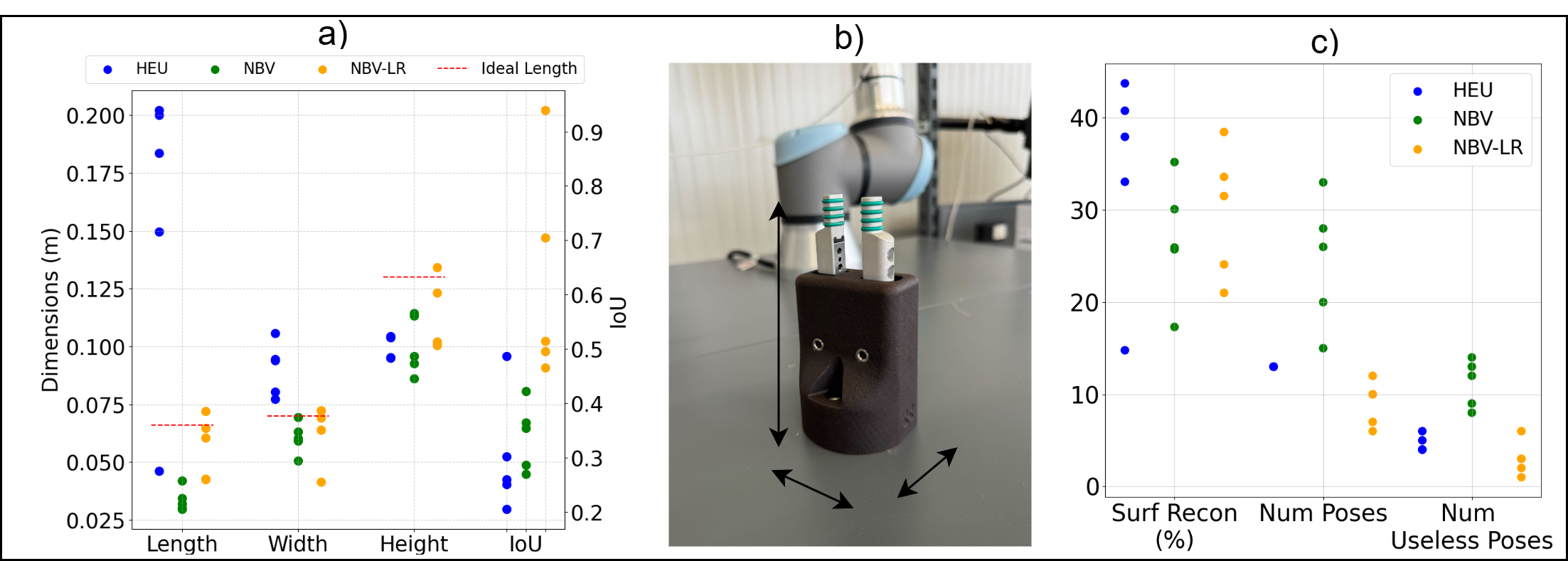}}  \\ 
{\hspace{-0.4cm}\includegraphics[width=0.5\textwidth]{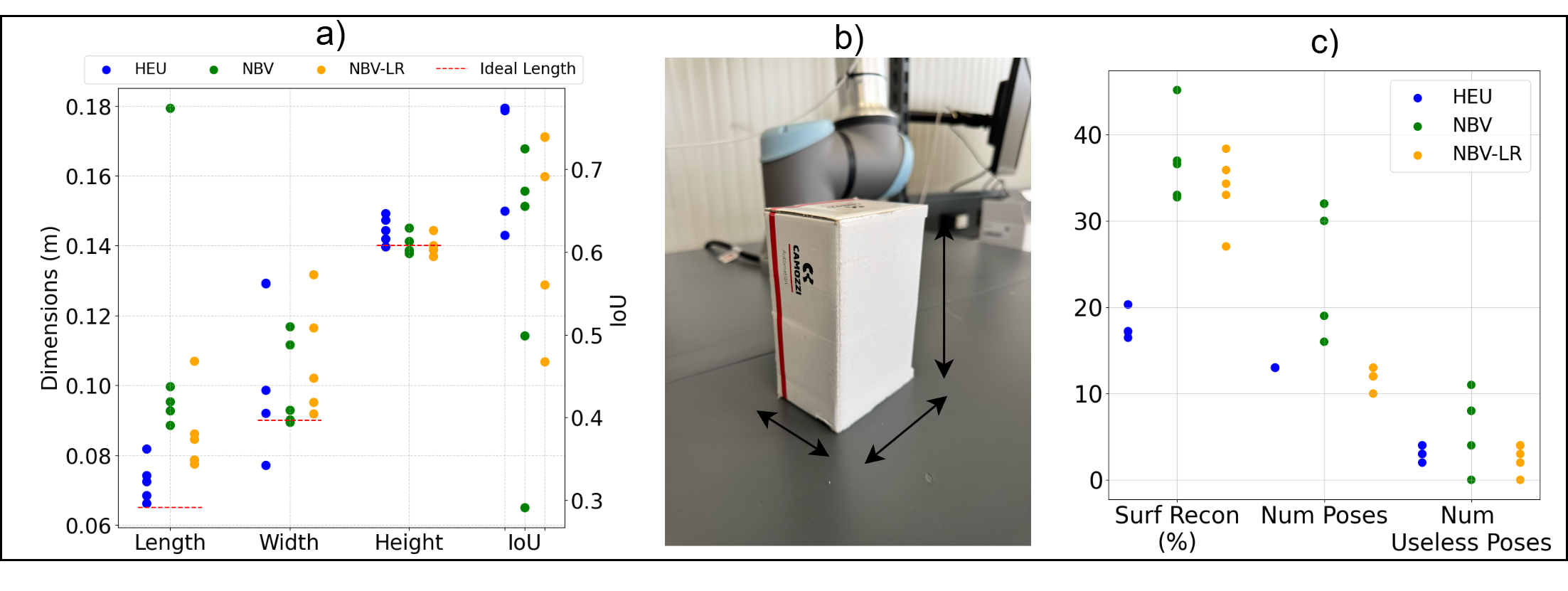}} & 
{\hspace{-0.1cm}\raisebox{0.17cm}{\includegraphics[width=0.5\textwidth]{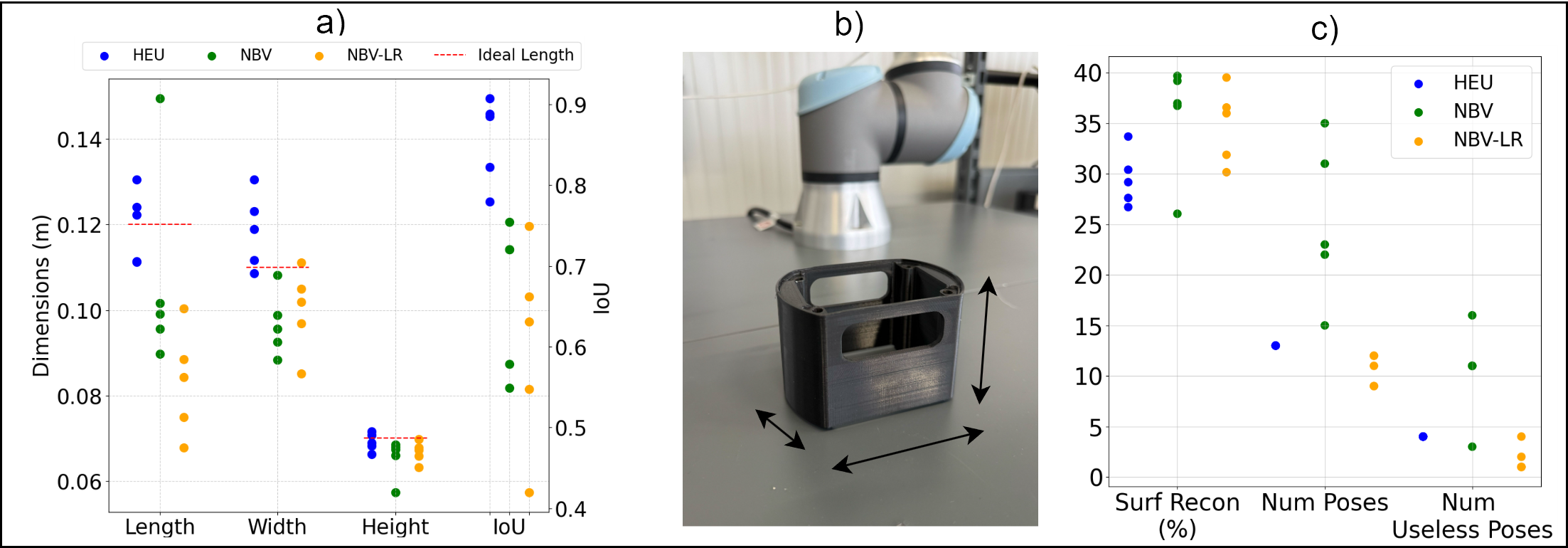}}} \\ 
\end{tabular}
\vspace{-0.3cm}
\caption{Acquisition results for four objects: the \emph{horizontal box} (top left), the \emph{vertical box} (bottom left), the \emph{gripper} (top right), and the \emph{flange} (bottom right). 
For each object:
(a) shows the comparison of estimated dimensions (length, width, height) and IoU, 
(b) displays the object with its canonical dimensions,
and (c) presents the surface reconstruction percentage, number of poses, and number of useless poses.}
\label{fig:all_experiments}
\end{center}
\end{figure*}

The \emph{Visited Gain \( G_v \)} prevents the algorithm from revisiting already explored regions. Following the approach presented by ~\cite{reference_paper}, a coarser discretization of the workspace is applied compared to the Octomap. Each discrete cell is assigned a value of 1 if already visited or 0 otherwise. The gain is then defined as:
\begin{equation} \label{eq:visited_gain}
G_v(x_T) = - \text{Visited}(x_T)
\end{equation}
The minus sign is used to reduce the overall gain when a cell has been already visited. 


\section{Experimental Validation}

The validation of this work compares the proposed \emph{LR-NBV} algorithm with two alternatives. The first alternative is a standard \emph{NBV} approach that follows the algorithm described in Algorithm~\ref{alg:nbv}, characterized by the same \emph{Exploration Gain} and \emph{Quality Gain} introduced in Section \ref{utility function}. Notably, this standard approach does not utilize the \emph{Density Gain} and \emph{Visited Gain}, also described in Section \ref{utility function}, which are key contributions of this work. This comparison aims to evaluate the effect of a utility function primarily based on exploration, as is common in most works in the literature. The second alternative is a heuristic method (\emph{HEU}) that requires the object position and generates 13 evenly spaced poses around an ellipse centered on the workspace. The experimental validation was conducted by performing 5 reconstructions for 4 different objects, shown in Fig. \ref{fig:all_experiments}, for the three proposed algorithms.


\subsection{Experimental Setup}

The algorithms were tested in a real-world setup using a UR5e robotic arm equipped with a VL53L8CX time-of-flight sensor mounted on the end effector. The sensor features a $45^\circ \times 45^\circ$ field of view (FoV) and an 8x8 resolution, making it a very low-resolution device. The tests were conducted on a Lenovo Yoga Pro 7i Gen 9 laptop with an Intel® Core™ i7-13700H processor and 32GB of RAM.
The implementation was based on Robot Operating System (ROS), utilizing the Octomap Library for dynamic measurement management and spatial representation, and MoveIt for motion planning. The parameters used during the experimental phase are summarized in Table~\ref{tb:parameters}.


\begin{table}[hb]
\begin{center}
\caption{Parameters used for the experiments.}\label{tb:parameters}
\begin{tabular}{lc}
\hline
\textbf{Parameter} & \textbf{Value} \\ \hline
\texttt{num\_iterations} & 300 \\
\texttt{sample\_per\_iteration} & 50 \\
\texttt{num\_cached\_nodes} & 50 \\
\texttt{visited\_cells\_grid\_size [m]} & 0.18 \\
\texttt{exploration\_weight} (\( w_e \)) & 500.0 \\
\texttt{density\_weight} (\( w_d \)) & 0.05 \\
\texttt{quality\_weight} (\( w_q \)) & 0.08 \\
\texttt{visited\_weight} (\( w_v \)) & 500.0 \\
\texttt{Octomap resolution [m]} & 0.01 \\ \hline
\end{tabular}
\end{center}
\end{table}

The evaluation of the algorithms was based on four metrics. \emph{1) Bounding box dimensions:} These represent the estimated dimensions of the object and are complemented by the intersection over union (IoU) index, which quantifies the precision of the reconstruction, which is critical for selecting appropriate-sized containers in packaging lines. \emph{2) Number of poses:} this serves as a proxy for time efficiency in robotic packing applications, as the execution time between iterations for NBV-based algorithms averages under 2 seconds on the experimental setup (excluding robot movement time). \emph{3) Surface coverage:} represents the percentage of the surface of the object reconstructed during the process. \emph{4) Number of useless poses:} refers to poses that do not increase the reconstructed surface and instead cause a decrease in the overall reconstruction quality. These metrics were extracted and are represented across five measurements for each object in Fig. \ref{fig:all_experiments}.

\subsection{Results} 

We conducted the Whitney U-test to compare the methods LR-NBV, NBV, and HEU across all metrics and experiments. The results indicate that LR-NBV achieves statistically significant improvements ($p < 0.05$) in terms of IoU, particularly for the objects \emph{horizontal box} and \emph{gripper}. For the other two objects (\emph{vertical box} and \emph{flange}), differences ($p < 0.05$) are observed for at least one dimension (length, width, or height), highlighting the ability of LR-NBV to provide more accurate bounding box reconstructions in several cases compared to the baselines. In terms of efficiency, LR-NBV consistently requires a significantly lower number of poses ($p < 0.05$) to complete the reconstruction, often less than half the poses required by NBV, while maintaining a reconstruction percentage that is at least comparable. This improved efficiency stems from the ability of LR-NBV to better prioritize informative poses, reducing redundant acquisitions. Conversely, NBV exhibits a higher number of \emph{useless poses}, which temporarily decrease the reconstruction percentage. These fluctuations extend the process as additional poses are required to recover the reconstruction quality lost due to these redundant acquisitions. However, even though we obtained satisfactory results for two of the four objects, the results for the remaining two objects show that there is still room for improvement. Therefore, more extensive evaluations will be conducted.

\section{Conclusions}


This work has demonstrated the potential of low-resolution sensors, characterized by low cost and ease of integration, for object reconstruction in robotic perception. The proposed algorithm has shown that sufficiently accurate reconstructions can be achieved to estimate object dimensions while significantly reducing the number of acquisitions compared to traditional methods. Future work will focus on validating the adequacy of the reconstructed surfaces for grasp planning and evaluating the algorithm performance on objects with more complex geometries using different sensors.

\begin{ack}
The authors acknowledge \emph{Camozzi Research Center} for providing sensor and robotic equipment, as well as for their support in this work.
\end{ack}

\bibliography{ifacconf}             
                                                   






\end{document}